\documentclass[twoside,11pt]{article}

\usepackage{blindtext}
\usepackage{booktabs}
\usepackage{jmlr2e}

\usepackage{lastpage}
\firstpageno{1}

\begin{document}

\title{Representational Alignment Supports Effective Machine Teaching}

% \author{\name Author One \email one@stat.washington.edu \\
%        \addr Department of Statistics\\
%        University of Washington\\
%        Seattle, WA 98195-4322, USA
%        \AND
%        \name Author Two \email two@cs.berkeley.edu \\
%        \addr Division of Computer Science\\
%        University of California\\
%        Berkeley, CA 94720-1776, USA}

% \editor{My editor}

\maketitle

\begin{abstract}%   <- trailing '%' for backward compatibility of .sty file

A good teacher should not only be knowledgeable; but should be able to communicate in a way that the student understands -- to share the student's representation of the world. In this work, we integrate insights from machine teaching and pragmatic communication with the burgeoning literature on representational alignment to characterize a utility curve defining a relationship between representational alignment and teacher capability for promoting student learning. To explore the characteristics of this utility curve, we design a supervised learning environment that disentangles representational alignment from teacher accuracy. We conduct extensive computational experiments with machines teaching machines, complemented by a series of experiments in which machines teach humans. Drawing on our findings that improved representational alignment with a student improves student learning outcomes (i.e., task accuracy), we design a \textit{classroom matching procedure} that assigns students to teachers based on the utility curve. If we are to design effective machine teachers, it is not enough to build teachers that are accurate -- we want teachers that can align, representationally, to their students too.

\end{abstract}


\section{Introduction}

% {\color{red}I think the overall argument we want to make in the first part of the intro is that expertise and representational alignment are both important contributing factors on teaching performance, but that contribution of representational alignment is modulated by classroom size and diversity.}

The proliferation of digital education resources and AI systems has enabled individual (human or machine) teachers to reach potentially millions of students. There appears to be growing interest in identifying top educators in each subject and have all students learn that subject from them. For example, Massive Open Online Courses (MOOCs) promised to revolutionize education by having educators from top research institutions record lectures in their domain of expertise and make course materials available online for any learner to interact with.
% What makes an effective teacher? Massive Open Online Courses (MOOCs) promised to revolutionize education by having educators from top research institutions record lectures in their domain of expertise and make course materials available online for any learner to interact with. 
However, after many trials, this expert-first approach to online learning proved not as generally effective and accessible as hoped for \citep{doi:10.1126/science.aav7958}, with courses delivered more traditionally by local teachers often showing better outcomes, in-person and online \citep{kelly2014disruptor}. 
More recently, powerful monolithic AI systems like ChatGPT have gained hundreds of millions of users, many of whom are using them, or the rapidly growing collection of educational applications they power, to learn new subjects.  
However, as these systems begin to outperform humans on some tasks \citep{Strachan2024theorymind, van2023clinical, Thirunavukarasu2024LLM}, their internal representations, at times, seem to become less human-like \citep{fel2022harmonizing, muttenthaler2022human}, setting up a tension between domain expertise and the ability to map knowledge into human-understandable spaces.
Similarly, many can relate to having teachers who are brilliant experts in their domains, but who struggle to instill knowledge in their students \citep{CARTER1987147,hinds2001bothered}. Yet many recent public education proposals focus primarily on increasing teachers' domain expertise and much AI research continues to primarily focus on improving the knowledge or expertise of the agents being developed. This begs the question, \textbf{what makes teaching effective, and is assigning all students to a single expert teacher the most effective strategy for improving student outcomes?} 

Understanding which factors contribute to effective teaching can help us determine optimal strategies for assigning students to teachers. While a teacher's domain expertise is certainly an important factor in determining student outcomes, we bridge ideas from the burgeoning subfield of \textit{representational alignment}~\citep{sucholutsky2023getting} with work on machine teaching, and the cognitive science of pedagogy, to propose that 1) representational alignment between teachers and their students, and 2) the size and diversity of the classroom, are likewise both key factors in determining the effectiveness of human and machine teaching. 

To test this hypothesis, we design a simple student-teacher cognitive task that %: \textbf{is it better to learn from a non-expert peer knows how you think or from a domain expert who does not?} 
enables the experimenter to independently control the teacher's expertise on the underlying task, and the degree to which their representations of the task are similar to the student's. In our first experiment, we investigate whether representation alignment plays a role in learning outcomes. 
Through simulations and a study where machines teach humans, we establish the relationship between teacher expertise, teacher-student alignment, and student performance. We find that representationally-aligned teachers with a high error rate on the underlying task can outperform highly accurate but representationally misaligned teachers. % that misaligned expert teachers may not be as valuable as peer teachers who are representationally aligned to their students. %This suggests the possibility that simply assigning all students to the teacher with the most expertise (e.g., a single powerful LLM, or a single MOOC instructor) may not be an optimal strategy.

\begin{figure}[h!]
    \centering
    \includegraphics[width=0.8\linewidth]{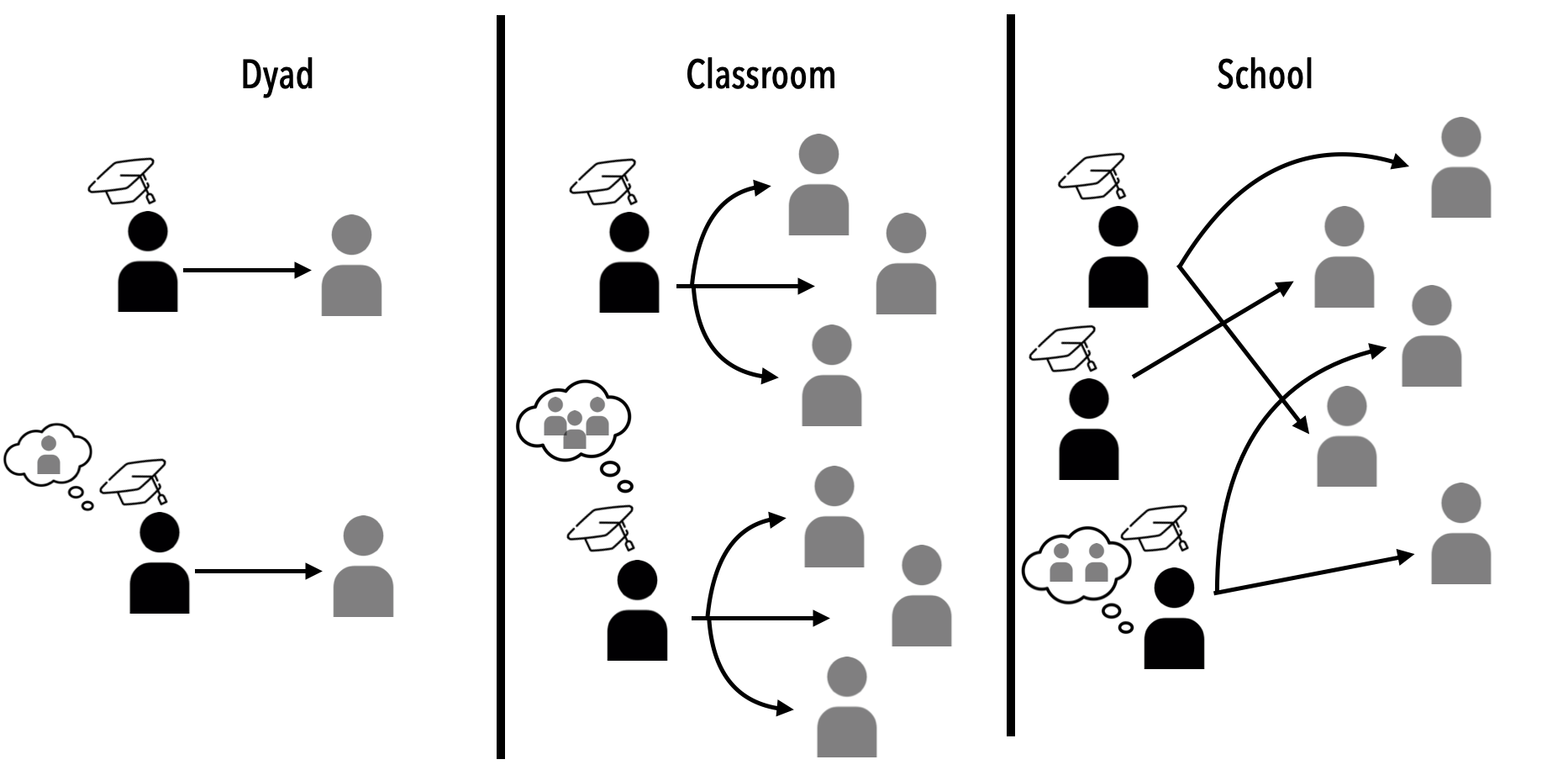}
    \caption{Schematic of various teaching set-ups. \textbf{(Left:)} Dyadic interaction between a single teacher and a single student. The teacher selects an example, or set of examples, to communicate to the student. Teachers can be ``student-centric'' and infer their student's representations when deciding which examples to select (left bottom). \textbf{Middle:} ```Classroom'' setting wherein a single teacher selects a set of examples to provide to all students in their ``class'' at once; student-centric teachers reason about all students in their classroom to make their selections (middle bottom). \textbf{Right:} ``School'' setting wherein many different teachers are matched with different students; each student is matched with a single teacher.}
    \label{fig:main-schematic}
\end{figure}

The results of the first experiment suggest that if a teacher can learn to adapt their representations to match the student, then the student's learning outcomes can potentially significantly improve. But what happens when this kind of ``student-centric'' teacher has to teach a whole class of students with individual difference in representations? 
In our second experiment, we investigate the role of classroom size in moderating the effect of representational alignment 
 on teaching outcomes. %People are diverse both in their learning preferences and their representations of the world [CITE Latent diversity paper], so no single teacher, human or machine, can be perfectly representationally aligned with all students. Instead, skilled human teachers can often infer their students' representations and adapt their teaching accordingly. 
We extend our task from a teacher interacting with individual students to interacting with a class of representationally diverse students where the material they present is simultaneously shared with all students in the class (e.g., like in a lecture; see Figure \ref{fig:main-schematic}). %We investigate whether adapting to individual differences in student representations is a viable teaching strategy (see Figure \ref{fig:main-schematic}).
We find that making teachers ``student-centered"  %who try to adapt to the students' representations outperform ``self-centered'' students who just teach based on their own representations, 
does lead to improved teaching outcomes, but the marginal improvement in student outcomes rapidly drops off as the class size or representational diversity increases. This suggests that adapting to individual differences in student representations may only be a worthwhile strategy in small-group settings and that simply assigning all students to a single expert teacher, even if they are student-centered, may still not be an optimal strategy for maximizing student outcomes. 

% Based on the finding then that -- at least in our setting where a teacher can only make a single aggregate selection of examples to provide to students -- there may not be a single, ``best'' teacher that can be assigned to all students, 
In our third experiment, we compare different strategies for assigning students to teachers. 
Leveraging the insights from the first two experiments, we design a preliminary \textit{classroom matching procedure} that takes into account representational alignment, teacher expertise, and class size when optimizing learning outcomes, and find that it outperforms both random assignment and MOOC-style assignment. 

Taken together, our study emphasizes the importance of considering student-teacher representation alignment -- not just teacher expertise -- in pedadogic settings, which is especially important if we want to design AI thought partners that can think with us and help us grow~\citep{collins2024building}. Our results also suggest that personalization, in particular ``student-centric'' teaching that accounts for individual differences in representation, may be crucial for improving learning outcomes, but is often infeasible in large classes. While it would be great to have domain experts teaching every subject, representationally-aligned peers, or personalized AI tutors, may need to fill the last-mile gap in education.

\section{A Controlled Domain for Single-Shot Teaching}%Simulations}
\label{simulations}

\subsection{Setup} 

\textbf{Task.} We construct a grid environment of size $N \times N$ where there are $K$ categories. Each grid cell is defined by its location within the grid with a label $k \in \{1..., K\}$. The student's goal is to correctly label each point. To support the student, the teacher can \textit{reveal} labels on the grid (i.e., creating teaching materials $L_0$ which consist of grid points and their associated labels). We consider two different label structures ($f$): one wherein there are $N$ classes (one class per column) and one with 4 classes (partitioning the grid into quadrants). Our experiments are reproducible and all the implementations for all computational experiments will be made available open-source upon publication.

\textbf{Representation.} The coordinates of grid cells form the student ($Y_s$) and teacher's ($X$) representations. We compute representational alignment as correlation between pairwise Euclidean distances. We corrupt the teacher's representation (in order to create misalignment between teacher space $X$ and student space $Y_s$) by sampling a number of grid cells to swap such that cell $(i,j)$ for the student appears as cell $(i',j')$ for the teacher (i.e., the mapping $T_s$ from the teacher to the student would reverse these swaps). 

\textbf{Student and teacher models} We instantiate our student ($g_s$) as a 1-nearest neighbor (1-NN) classifier, who takes as input the teacher's revealed examples ($L_s$) and classifies each of the unlabeled points. Student performance ($V_s$) is computed as the accuracy of their classifications over the unlabeled points. The teacher chooses $K'$ points intended to maximally help the student (whom the teacher ``knows'' is a 1-NN classifier) to achieve high accuracy on the remaining points. We assume the teacher has access to labels for all cells; however, the ``erroneous'' teacher with some probability assumes the wrong label on a cell (i.e., $f'$ is different from $f$, which can ripple into their selections accordingly). The teacher computes the centroid of each class (using its own believed labels $f'$, which may have errors) and selects one example per class to reveal to the student. The teacher reveals its believed labels to the student for the selected points. %As such, the teacher selects the centroid from each (believed) class. 
 
% \textbf{Learning rule}

% \begin{figure}
%     \centering
%     \includegraphics[width=0.5\linewidth]{example-image-a}
%     \caption{Simulation results }
%     \label{fig:sim-res}
% \end{figure}

% \begin{figure}%
%     \centering
%     % \subfloat[\centering 5x5 grid, 5 class]{{\includegraphics[width=5cm]{figures/5x5_5cls.png} }}%
%     % \qquad
%     \subfloat[\centering]{{\includegraphics[width=5cm]{figures/10x10_10cls.png} }}%
%         \qquad
%         \subfloat[]\centering]{{\includegraphics[width=5cm]{figures/repr_alignment_diff.png }}
%     \caption{PLACEHOLDER -- simulation results \textit{which viz is better...}. In the heatmap, color indicates student performance (accuracy on unlabeled points); higher (brighter colors) is better. 10x10 grid; 10 class.}%
%     \label{fig:sim-res}%
% \end{figure}

\begin{figure}[htbp]
    \centering
    \begin{subfigure}
        \centering
        \includegraphics[width=0.3\textwidth]{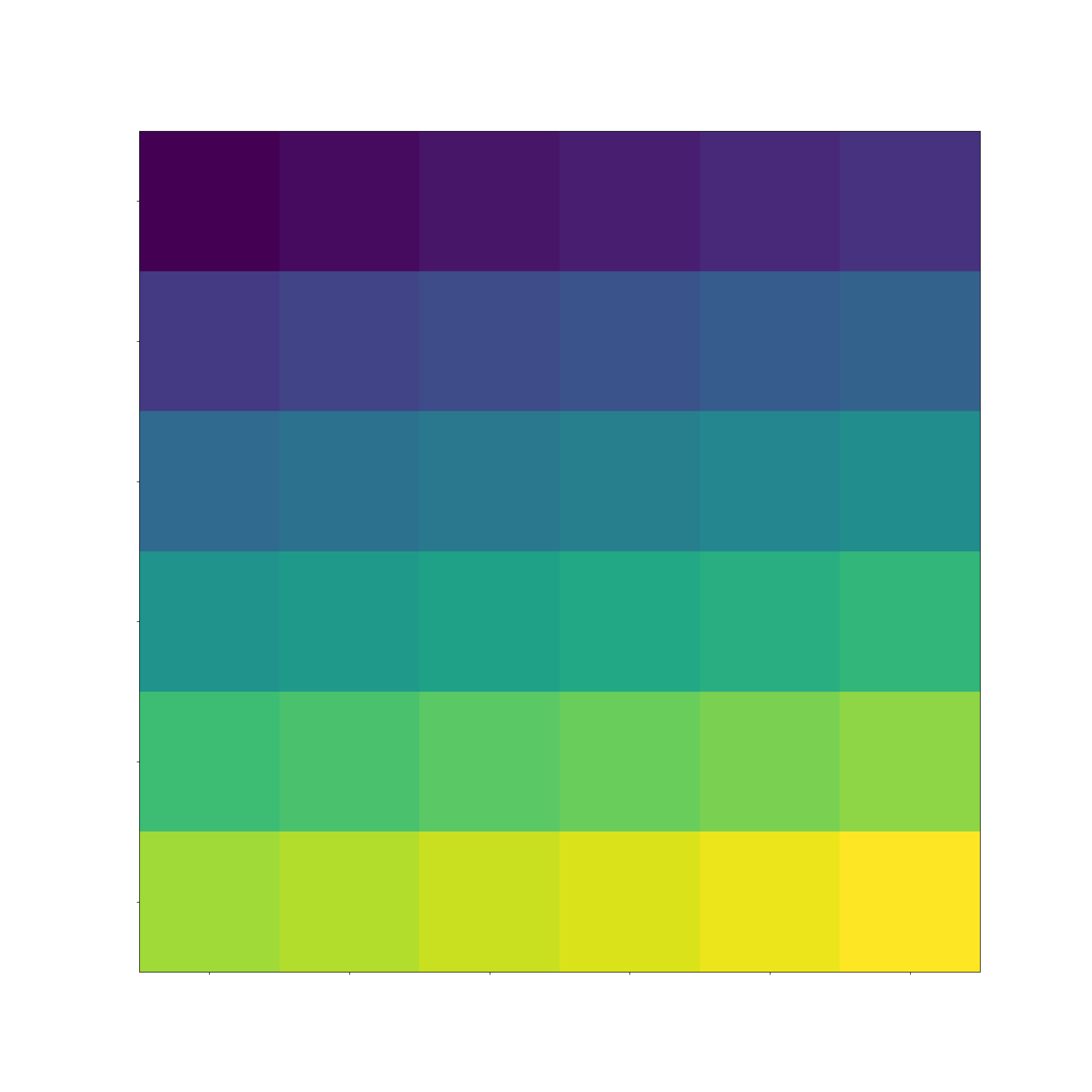} % Replace with your image file
    \end{subfigure}
    \hfill
    \begin{subfigure}
        \centering
        \includegraphics[width=0.3\textwidth]{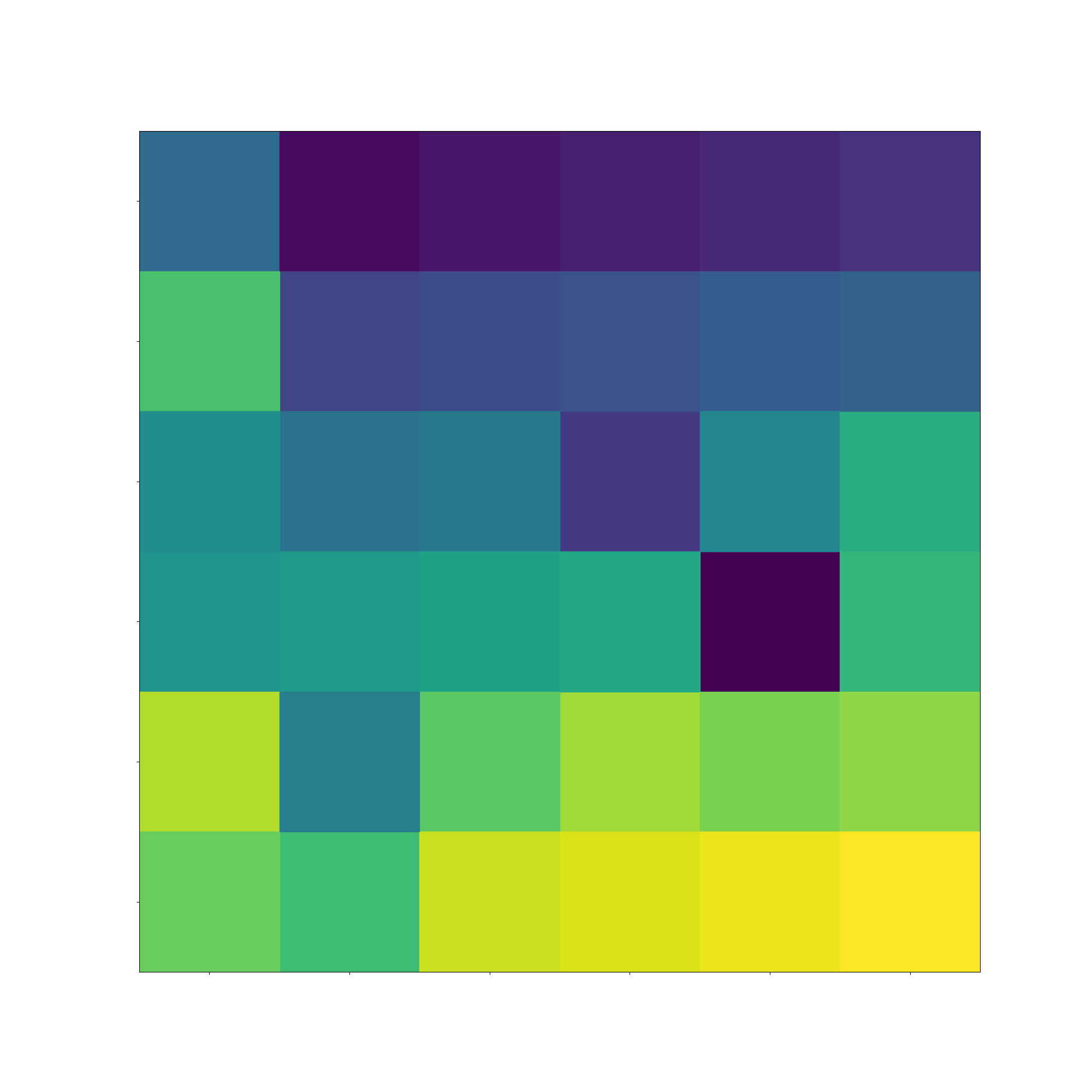} % Replace with your image file
    \end{subfigure}
    \hfill
    \begin{subfigure}
        \centering
        \includegraphics[width=0.35\textwidth]{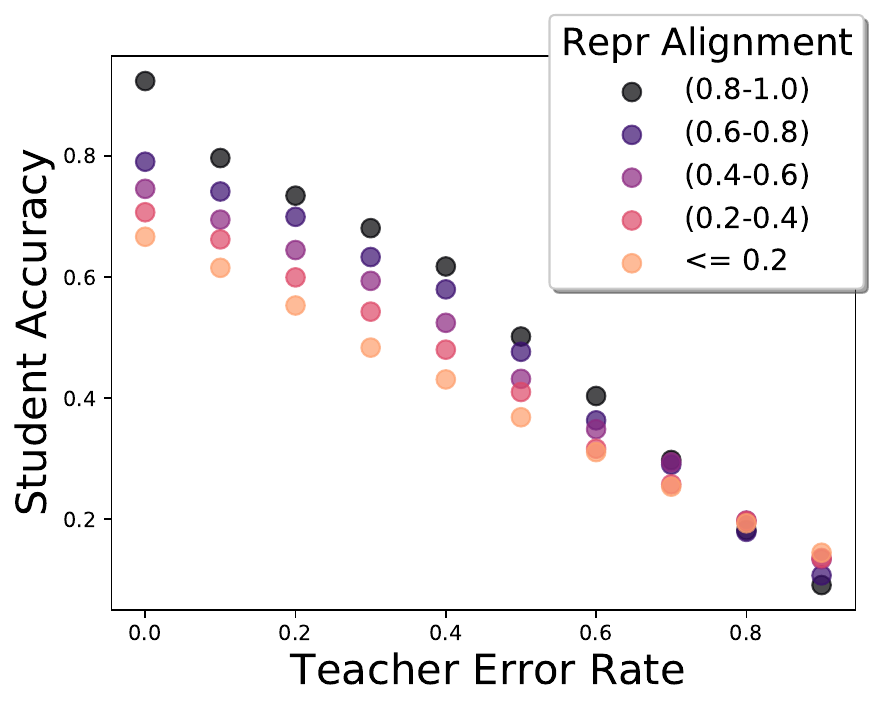} 
    \end{subfigure}
    \caption{Representational (mis) alignment between students and teachers can impact student learning outcomes. \textbf{(Left:)} Sample student representation. \textbf{(Middle:)}: A misaligned teacher. Color indicates the identity of the stimuli as a visual aid to illustrate different representations of the same set of stimuli. The same stimuli are indicated with the same color between student and teacher. \textbf{(Right:)} Tracing out a utility curve between teacher error rate and representational alignment between teacher-student in simulations (where representations are grid locations, as in the left and middle plots). Points are colored by representational alignment (higher/darker is more aligned). Student learning outcomes are operationalized as accuracy at classifying grid cells (higher is better).}
    \label{fig:sim-res}
\end{figure}

\subsection{Relationship between representational alignment and teacher performance}

In Figure \ref{fig:sim-res}, we trace out a relationship between student-teacher representational alignment, teacher error rate, and student accuracy. We uncover instances wherein students can achieve higher performance by learning from teachers who are \textit{more erroneous} (``less expert'') provided the teachers are representationally aligned with the students than comparatively more expert but representationally misaligned teachers, underscoring that it is not just the accuracy of a teacher that matters for student learning outcomes. However, we note that there is a tradeoff: more representationally aligned teachers are not always better if the teacher is too noisy or misguided in their labels. Yet, for a fixed teacher error rate, higher representational alignment is always better for a student (provided the error rate is not too high). We uncover similar curves across grid sizes and the number of categories (see Appendix~\ref{dyadic-utility-curve}). 

% We see that across varied grid sizes and number of categories [show diff grids?], for a fixed teacher accuracy level, the student is able to achieve higher performance when the teacher selecting examples is more representationally to them. However, we note that teachers who have high error rate do not help students acheive high performance, even if representationally aligned; likewise, students can achieve good performance with ... 

% [todo: more emphasis on interesting case of a student acheiving higher acc from a teacher who is more repr aligned but less accurate -- in Figure\ref{fig:sim-res} (right) we see that students can achieve higher performance with teachers who are less ``expert'' (higher error rate) [blue] if they are more representationally aligned than ``expert'' teachers [red]] 

% \subsection{Impact of algorithmic alignment}

% We next consider a setting where the student's classification rule [match notation from above] differs from what the teacher ``believes'' the student's rule is. The teacher and student may have the same representation of the world (cell locations match), but if the teacher believes the student operates over those representations differently (i.e., that they are more or less advanced), the teacher may select less appropriate examples and student performance again suffers. Here, we vary the student's algorithm by varying $k$ in the student's nearest neighbor classification [ or vary alg?]. 

\section{Validating the Relationship between Representation Alignmnet and Teaching Expertise with Humans}%Experiments with Machines Teaching Humans}
\label{machine-human-exps}

\subsection{Setup}

\textbf{Participants.} We recruit 480 participants from Prolific~\citep{palan2018prolific}. We filtered the participant pool by country of residence (United States), number of completed studies ($>100$), and approval rate ($>95\%$).  Participants gave informed consent under an approved IRB protocol.

\textbf{Task.}
We design a task for our machines to teach humans about categories, in which participants see a grid of stimuli; for each cell in the grid, there is an underlying true category. Our simulated teacher model selects labels based on these underlying categories, and participants see these labels with the grid. Participants must then categorize all the unlabeled stimuli on the grid using the teacher’s labels. We do not inform participants of the number of examples per category.
We investigate two structures of categories: one class per column of the grid (``cols’’) and one class per quadrant of the grid (``quad’’). Note that these categories induce labeling functions that the students \textit{should} be able to learn; they are tractable (column structure and block structure).
There were two different stimuli sets. The first (\textit{simple-features}) is the closest analog to our simulated experiments, in which participants saw a $6 \times 6$ grid with blank cells, so the features are completely expressed via the coordinates of the grid. The second (\textit{salient-dinos}) is a more rich set of stimuli, wherein participants see a $7 \times 7$ grid of dinosaur (``dino’’) images from \citep{malaviya2022can}. Dino stimuli were defined by nine different features (e.g., body length, neck length, neck angle) and organized on the grid by two principle components of those underlying features. For a visualization of the participant's view, see Figure \ref{fig:experiment-dino-grid}.
For each condition, different teachers were generated from our model, sampling across varying levels of alignment. This structure leads to 24 different conditions (2 stimuli sets $\times$ 2 category structures $\times$ 6 teacher alignment levels) for which we collect 20 participants each.

\textbf{Models and Example Selection.} We employ the same model types as in our simulations. Teachers are self-centered and assume that students are 1-NN classifiers\footnote{We acknowledge such an assumption is highly simplistic for students and encourage future work to explore alternate models of students.}. In both settings, we assume the representations of teachers and students can be expressed through their two-dimensional grid locations. For the simple-grid setting, there are no features for the human to use for their categorization beyond grid cell location; and in the \textit{salient-dinos} setting, features were defined by two principal components (which we can use as grid coordinates). We again induce representation misalignment between teacher and student by shuffling the stimuli on the grid. We sample a set of teachers spanning a range of representational alignments. We select a single set of points for each teacher, assuming the teacher has perfect accuracy. We explore alternate labeling functions to simulate alternate teacher error rates post-hoc (see Appendix~\ref{human-exp-details-appendix}). %; however, offers a natural direct analogue from our simulations as we embark on a first exploration of machines teaching human. 

%In both cases, the teacher assumes the student is a 1-NN classifier. 

%We note that this assumption is inherently limited but offers a sensible [... support from cogsci? note as limitation?] 

%We consider two classes of models and representation corruptions. For the \textbf{simple-features} task, we employ a rational pragmatic teacher, directly analogous to our simulated experiments. Corruptions for both tasks are again pairwise swap. 
%In our \textbf{salient-dinos} task, we consider a different kind of representation misalignment: \textit{projections}; i.e., the teacher only sees a subset of the full nine features. 
%In both cases, the teacher assumes the student is a 1-NN classifier. We note that this assumption is inherently limited but offers a sensible [... support from cogsci? note as limitation?] %We sample [...number...] of models from each bin of representational alignment and modulate error rate ... 

\subsection{Results}
\begin{figure}
    \centering
    \includegraphics[width=0.49\linewidth]{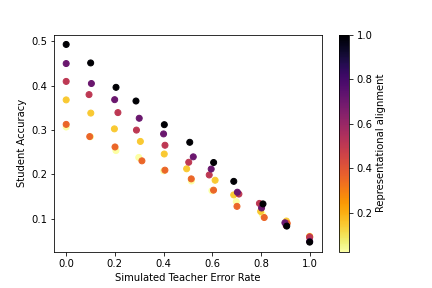}
    \includegraphics[width=0.49\linewidth]{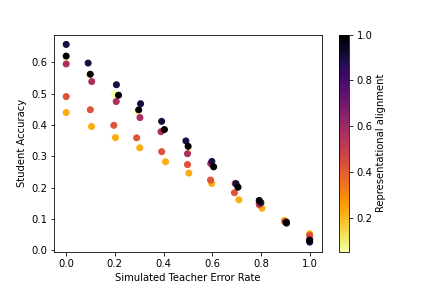}
    \includegraphics[width=0.49\linewidth]{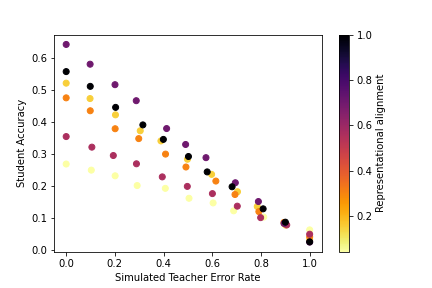}
    \includegraphics[width=0.49\linewidth]{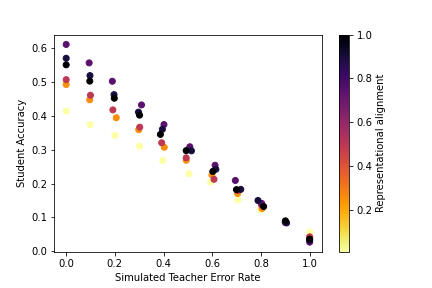}
    \caption{Tracing out a utility curve between teacher error rate and representational alignment between machine teachers and human students. Points are colored by representational alignment (higher/darker is more aligned). Student learning outcomes are operationalized as classification accuracy (higher is better). \textbf{(Left:)} Results when classifying grid points with no images (the \textit{simple-features} task). \textbf{(Right:)} Results when classifying dino images (the \textit{salient-dinos} task). \textbf{(Top:)} Results with one class per quadrant. \textbf{(Bottom:)} Results with one class per column (6 for simple-features, 7 for salient-dinos).}
    \label{fig:human-res-main}
\end{figure}

\textbf{Representational alignment between machine teachers and human students can affect learning outcomes.}
We construct a utility curve paralleling our simulations by post-hoc varying teacher error rate (see Appendix~\ref{further-analyses}). We find in Figure \ref{fig:human-res-main} that across both tasks, generally, higher representational alignment induces higher average student accuracy, and report correlations in Table~\ref{tab:human-results}. By computing a simple linear regression on the points with no teacher error, we find that a 0.1 increase in representational alignment corresponds to a 2.1\% increase in student accuracy for the simple-features conditions ($r=0.59, p=0.054$) and a 1.4\% increase in student accuracy for the salient-dinos conditions ($r=0.63, p=0.037$).  

Thus, we find that even large increases in teacher error rate can be offset by increasing representational alignment (e.g., a teacher with error rate $0$ and representational alignment of $0.3$, has similar student outcomes as a teacher with error rate $0.4$ and representational alignment $0.8$). 
However, we note that the ordering of high representational alignment is less clear, particularly for the settings where each class corresponds to a column. We posit that people have a strong prior against classes being distributed as columns, and find that especially for the columns conditions, participants would often label using strategies that did not correspond to nearest neighbor classification (e.g., several participants labeled in a way that corresponded to different types of tilings). This motivates future work extending our framework to account not only for representational alignment between teachers and students but also for computational or algorithmic alignment.

\section{Classroom Matching}
\label{classroom-matching}

We have demonstrated that a teacher's representational alignment with a student matters alongside teacher accuracy for student outcomes. Yet, in real teaching settings, rarely is a teacher matched with a single student; rather, they may need to select examples that suit a cohort, or group, of students at once. Selecting such examples grows all the more challenging when students have different representations. When provided with a ``school'' of teachers and students, how can we simultaneously construct student groupings and which teacher to allot for which group of students? We begin to explore this question through a series of ``classroom matching'' experiments. In particular, we develop a \textit{classroom matching procedure} which, given a pool of students and teachers, pairs groups of students to teachers based on our representational alignment-teacher utility curve. But, some students may still be left behind. To begin to address lower-performing students in our pool who may not have a representationally close teacher, we depart from our self-centered teacher set-up to consider a \textit{student-centric} teacher: one who is able to select examples in recognition of the students' representations of the problem. We find that we can get a boost of up to an extra 10\% of accuracy points for a subset of students who are representationally distinct from the rest of the pool of students and teachers -- crucially, provided there are not too many students grouped into the classroom. We emphasize though, that our analogy to ``classrooms'' is explored in simulation with machines teaching machines; substantial future work is needed -- and motivated by this work -- to cache out possible links between representational alignment, teacher error rate, and classroom properties like classroom size in practice.

\subsection{Setup}
% todo: revise for both cases

\textbf{Student and teacher pools.}
We focus on our \textit{simple-features} setting and extend our dyad (single teacher, single student) setting to simulated \textit{pools} of teachers and students over our same $6 \times 6$ grid. Here, we consider row-based labels (a new $f$). We include generalization experiments to the \textit{salient-dinos} setting in Appendix ~\ref{dinos-matching generalization}. We explore two different pools of students and teachers: (i) unstructured pools spanning a range of representational alignments and error rates, and (ii) clustered sets of students and teachers. For the latter, we construct a generative model over student-teacher populations wherein we have a set of clusters, with a fixed number of students per cluster share similar representations. We sample one similar teacher from each cluster with some error rate (sampled from a uniform distribution over $0$ to $0.5$). We then deliberately downsample from the available teachers to simulate the case where some students are representationally distinct from the other students and available pool of teachers. Additional details are included in Appendix \ref{classroom-pools}. For each experiment, we sample $10$ different teacher pools. 

\textbf{Matching procedures.}
We propose to match students using our utility curve to estimate their accuracy. That is, we compute the representational alignment between a student and teacher and index into a bucketed version of the utility curve\footnote{We recompute the curve by also averaging over samples of corrupted students as our first utility curve was constructed for dyad setting wherein the student's grid was never corrupted (see Appendix \ref{classroom-utility-curve}).} that we construct in Section \ref{simulations} using both the representational alignment and teacher's expected error rate (which we assume we have access to). The resulting performance provides an expected performance that we anticipate the student may attain if paired in a classroom with that teacher. We iterate over all teachers for each student and select the teacher who helps the student achieve the highest expected performance. We refer to this method as \textbf{Ours}. 

Additionally, we consider three baselines: (i) \textbf{Random} matching of students to teachers, (ii) \textbf{MOOC} which matches all students to the lowest error rate teacher to simulate the case non-local expert teacher arrangements, and (iii) \textbf{Optimal} wherein we match students to the highest attainable accuracy, giving an indication of the upper limit of performance that a matching algorithm could achieve. Importantly, gaps between (ii) and (iii) further drives home the importance of more than just teacher accuracy for informing student outcomes.

\subsection{Results}

\begin{table}[]
\centering
\caption{Student learning outcomes (accuracy) from different classroom matching approaches in the structured (clustered) pool setting. We have 10 representationally distinct clusters, each with 50 students, and sample 5 available teachers across the clusters (i.e., 50\% of the clusters will \textit{not} have a representationally aligned teacher available).}
\begin{tabular}{@{}lcccc@{}}
\toprule
Method  & Avg Acc & Bottom 10\% & Top 10\% & Pass Rate \\ 
\midrule
Random  & 0.33 $\pm$ 0.01 & 0.2 $\pm$ 0.02 & 0.49 $\pm$ 0.02 & 0.12 $\pm$ 0.03 \\
MOOC    & 0.37 $\pm$ 0.02 & 0.26 $\pm$ 0.02 & 0.54 $\pm$ 0.06 &  0.17 $\pm$ 0.07 \\
Ours    & \textbf{0.39 $\pm$ 0.02} & \textbf{0.27 $\pm$ 0.02} & \textbf{0.57 $\pm$ 0.06} & \textbf{0.23 $\pm$ 0.06} \\
\hline
Optimal & 0.43 $\pm$ 0.02 & 0.32 $\pm$ 0.02 & 0.6 $\pm$ 0.05 & 0.30 $\pm$ 0.06 \\
\bottomrule
\end{tabular}

\label{tab:classroom-matching-clustered}
\end{table}

\textbf{Putting our utility curve to work to design ``classrooms''.} %improves aggregate student outcomes over non-local experts}
We find that our matching algorithm, which groups students to teachers based on their representational alignment and teacher error rate, suggests that prudent student-teacher pairings are better than random matching and, particularly for top-performing students, are better than having assigned the student to an expert (minimal error rate teacher; MOOC) who may be representationally distinct (see Tables \ref{tab:classroom-matching} and \ref{tab:classroom-matching-clustered}). This observation is intriguing -- students may not achieve their full potential when paired with a representationally misaligned teacher, even if that teacher is an expert. We note that in the structured case (Table \ref{tab:classroom-matching-clustered}), wherein students and teachers have distinct clusters of representational alignment, assigning all students to the most accurate teacher (MOOC) does not do substantially better than random matching across some of the metrics. We observe performance gains for our utility curve-based matching across both pool types. However, we note that we do not yet attain optimal matching performance; such discrepancy can be attributed to a mismatch in our utility curve. %We posit that we may be able to boost 

\textbf{Student-centric teachers can cover students who are representationally ``left behind''.}
Even with our matching approach, students can be left behind if there is no suitable representationally aligned teacher for them. Can we better train teachers by drawing on our findings that representational alignment appears to be a key factor in student learning outcomes? Thus far, our teachers have all been self-centered; that is, they have a single representation that they operate over to select points. This setting may be realistic in practice for machine teachers that are not equipped with the ability to adapt to the students at hand. However, real human teachers -- and the machine teachers we may want to design -- ought to be able to adapt to the student at hand. We next consider \textit{student-centric} teachers that select points to maximize the average performance of the students in their pool (see Appendix \ref{student-centric-teacher-details}). 

We begin to explore student-centric teachers by appending a second stage to our matching procedure. After matching using our utility curve (as noted above), we greedily attempt to pair the lowest performing students with a student-centric teacher who chooses points by optimizing for the students in their pool (i.e., taking the students' representations into account). We continue incorporating the next lowest-performing students into the student-centric teacher's classroom until a student's attained accuracy with the original pairing is not improved by the student-centric teacher. We apply our procedure to the clustered pool structure noted above and find that it is beneficial to continue adding students up to a point: if the teacher is an expert (zero error rate), we can add all students from one cluster before we see detrimental performance across the pool of students assigned to said teacher. As the student-centric teacher's error rate increases, fewer students can be pooled before performance dropoff (see Appendix Figure \ref{fig:student-centric-greedy}).

%\textbf{Classroom size matters}
These results indicate the student-centric teachers can cover students who are representationally distinct and help boost their learning outcomes. However, classroom size matters,corroborating prior works in machine and human teaching~\citep{frank2014, yeo2019iterative, teacherYuzhe2018, zhu2018overview}. We conduct a deeper dive into the relationship between classroom size and student outcomes in our setting when student-centric teachers are available in Appendix~\ref{cls-size-student-centric}. Herein, we see that teachers who may try to overalign to all students at once in a large classroom induce poorer outcomes for the classroom writ large.

\section{Discussion and Limitations}
\label{discussion-and-limitations}

We have begun to characterize qualities of effective machine teaching through the lens of representational alignment. Our computational and human experiments elucidate that teacher accuracy is not all that matters for student learning outcomes: we also ought to care about a teacher's representational alignment with a student, a message with resonance to learning sciences literature. 

Our results herald practical implications. For machines to be effective thought partners who help us grow and learn, we may want them to be able to successfully model our representations; but, humans vary in our representations of the world \citep{peterson2018evaluating, marti2023latent}. Our classroom matching studies motivate designing for diverse pools of teachers if you want the best outcomes for diverse pools of students. ``Representations should be representative'' -- i.e., distribution of teacher representations should be representative of the students with care taken to classroom size. Further, our results link back to some of the challenges of deploying uniform curricular materials like MOOCs; even if machine teachers are superhuman experts in some respects, some students may benefit from less expert teachers who are more representationally aligned. We emphasize that our results in no way suggest that the MOOC approach of having top educators design teaching materials is somehow bad, but rather that some students will be better served by instead being matched with representationally aligned teachers, even if those teachers have less domain expertise. Appropriately matching students to teachers may become all the more important as the space of available AI systems expands rapidly (we already see a suite of large language models on offer for widespread use \citep{openai2023gpt4, touvron2023llama, jiang2023others, palmModel, pardos2024chatgpt}).

Yet, we emphasize that our work is a first step in the study of the relationship between representational alignment, teacher efficacy, and student-teacher matching. Our simulations always assume that students are 1-NN classifiers, which grossly undercuts the richness of human behavior. Further, we always assume that the student's representations are fixed; in practice, students may adapt their representations of the world over time, and it may be advisable for them to do so. Additionally, we consider only single-lesson settings and settings wherein students have no indication of the reliability of the teacher. We plan to extend our simulations and human experiments to multi-lesson sequences wherein students get feedback on teaching accuracy in the future. 

\section{Conclusion}

Expertise on a task is not sufficient to be a good teacher. Representational alignment between a teacher and student matters too. We trace out a utility curve between teacher accuracy, teacher-student representational alignment, and student accuracy to characterize the crucial relationship between representational (mis)alignment and student learning outcomes. We find that the most expert teacher is not always the most optimal and that the utility of representation is approximately 2\% (absolute) improvement in student accuracy for every 10\% (absolute) increase in representational alignment. This empirical result from a machine teaching lens, underscores the importance of having teachers, human or machine, capable of representing a diversity of students, or facilitating pools of teachers such that there is at least one teacher that any student can effectively learn from. Our work motivates further investigation into the relationship between human-human representational alignment and pedagogical effectiveness, not just for teacher-student interactions but peer-to-peer learning.

% % Acknowledgements and Disclosure of Funding should go at the end, before appendices and references

% \acks{All acknowledgements go at the end of the paper before appendices and references.
% Moreover, you are required to declare funding (financial activities supporting the
% submitted work) and competing interests (related financial activities outside the submitted work).
% More information about this disclosure can be found on the JMLR website.}


\bibliography{main}

\newpage

\appendix

\section{Broader Impact and Societal Risks}
\label{broader-impact}

As we discuss in Section \ref{discussion-and-limitations}, our work portends broader implications for the design of machine-human teaching setups where machines are intentionally built with representation alignment in mind, as well as representation diversity to safeguard against threats to inclusivity. It is possible that our simulations could support interventions in real classrooms, e.g., informing classroom size decisions drawing on measures of the representational diversity of a classroom pool and the expertise of the teacher. However, we heed caution in over-generalizing our results to settings where real student experiences and learning potential is at stake. Broad-brush application of AI systems in education has not been met with universal success \cite{reich2020failure} 
and inappropriately incorporated can have unintended impacts on student success \cite{doi:10.3102/0091732X20903304}. 

\section{Related Work}

\textbf{Learning Sciences.} The extended learning sciences community has studied aspects of what makes for a good teacher or computer-based teaching system. The expertise or quality of the teacher with respect to excellence of schooling, certification, and a teacher's own test scores have been observed to positively affect student learning \cite{rice2003teacher}. More classroom-adaptive qualities, like a teacher’s amount of experience in classrooms and teaching strategies employed (i.e., pedagogy) are also top contributing attributes \cite{rice2003teacher}. Closeness of representation to students with respect to demographic features has been shown to lead to more effective student performance \cite{dee2004teachers}, in part due to the role model effect, but also because teachers closer in these dimensions can serve as sociocultural interlocutors, helping translate the relevance of  material to students \cite{EGALITE201544, harfitt2018role}. Intelligent Tutoring Systems \cite{anderson1985intelligent}, growing out of the cognitive and learning sciences, have been a consistently effective paradigm of computer-based teaching \cite{wang2023examining}, primarily utilizing the pedagogy of mastery learning \cite{bloom19842}. They adapt the amount of prescribed practice based on a representation of the student’s level of mastery of the skill being worked on and provide procedural remediation in the problem-solving context. In a two-year, large-scale evaluation, a commercial ITS was found to be effective overall, but only demonstrated superior learning gains to standard classroom instruction in the second year. It was hypothesized that this may have been due to teachers needing to learn how best to align their classroom to the technology \cite{pane2014effectiveness}. 
%TODO: @Zach bring in some literature on machines teaching humans, tutoring systems (which model student capabilities not representations, don't adapt to student representations), knowledge tracing, the current gap (e.g. teaching a classroom or cohort of students). 21st century skills, dimensionality of STEM. Optional: how are curricula/teaching materials currently designed, how does classroom matching work, existing evidence of student outcomes differences with local vs non-local or peer vs expert teaching, 
% - ITS has a student model, but largely is an assessment of competencies 
% - ITS teaches procedures, not dimensionality expansion (maybe for discussion)
% - 

\textbf{Machine teaching.} Machine teaching aims to study the problem of teaching efficiency by characterizing such efficiency as the minimal number of effective data examples that is needed for a learner to learn some target concept. It has an intrinsic connection to optimal education~\citep{zhu2015machine}, curriculum learning~\citep{liu2017iterative,korakakis2023minimax} and optimal control~\citep{lessard2019optimal}. Depending on the type of learner, machine teaching can be performed in a batch setting~\cite{zhu2015machine,zhu2018overview,liu2016teaching} or an iterative setting~\citep{liu2017iterative,liu2018towards,liu2021iterative,qiu2023iterative,zhang2023nonparametric}. The batch teaching aims to find a training dataset of minimal size such that a learner can learn a target concept based on this minimal dataset. The iterative teaching seeks a sequence of data such that the learner can sequentially learn the target concept within minimal iterations. Complementary to these works, our findings indicate that, alongside the quality of examples that the teacher selects, it is also critical for both the teacher and the student to share similar representations.

\textbf{Pragmatic communication.}
Successful communication rests on our ability to understand others' beliefs and intentions~\cite{gweon2021inferential, velez2023teachers}. Indeed, even young children are sensitive to others' knowledge and competence when teaching \cite{liszkowski2008twelve, bridgers2020young} and learning \cite{bass2022effects, csibra2009natural, bonawitz2011double} from others. 
Inspired by Gricean pragmatics~\cite{grice1975logic}, recent computational models have formalized this process as recursive reasoning about others' latent mental states~\cite{chen2022hierarchical, goodman2016pragmatic, shafto2014rational}. Such pragmatic models have been used to facilitate human-AI interaction ~\cite{sumers2021learning, sumers2022talk, lin2022inferring, andreas2016reasoning, dale1995computational, fried2018speaker, wang2016learning, wang2020mathematical, ho2016showing, zhixuan2024pragmatic}.
Crucially, however, when either party \emph{fails} to accurately model the other's beliefs or perspective, human-human~\cite{aboody2023naive, sumers2023show} and human-AI~\cite{milli2020literal, sumers2022talk} communication can be significantly degraded. Our work adds to this literature by formalizing and analyzing the effect of \emph{representational} misalignment on communication.

\textbf{Representational alignment.}
Representational alignment~\citep{sucholutsky2023getting} offers a conceptual and grounded mathematical framework for characterizing teaching settings wherein two or more agents engage on some task. Already, ideas from representational alignment are providing new ways of thinking about machine learning efficiency ~\citep{sucholutsky2023alignment}, value alignment ~\citep{rane2023concept, wynn2023learning}, disagreement~\citep{oktar2023dimensions}, and applications like human \& machine translation~\citep{niedermann_sucholutsky_marjieh_celen_griffiths_jacoby_van_rijn_2024}. In this study, we show that representational alignment is a key dimension in predicting and optimizing student outcomes, with similar importance as the teacher's subject expertise.

\section{Additional Details on Problem Setting}
\label{theory}

\begin{figure}
    \centering
    \includegraphics[width=0.8\linewidth]{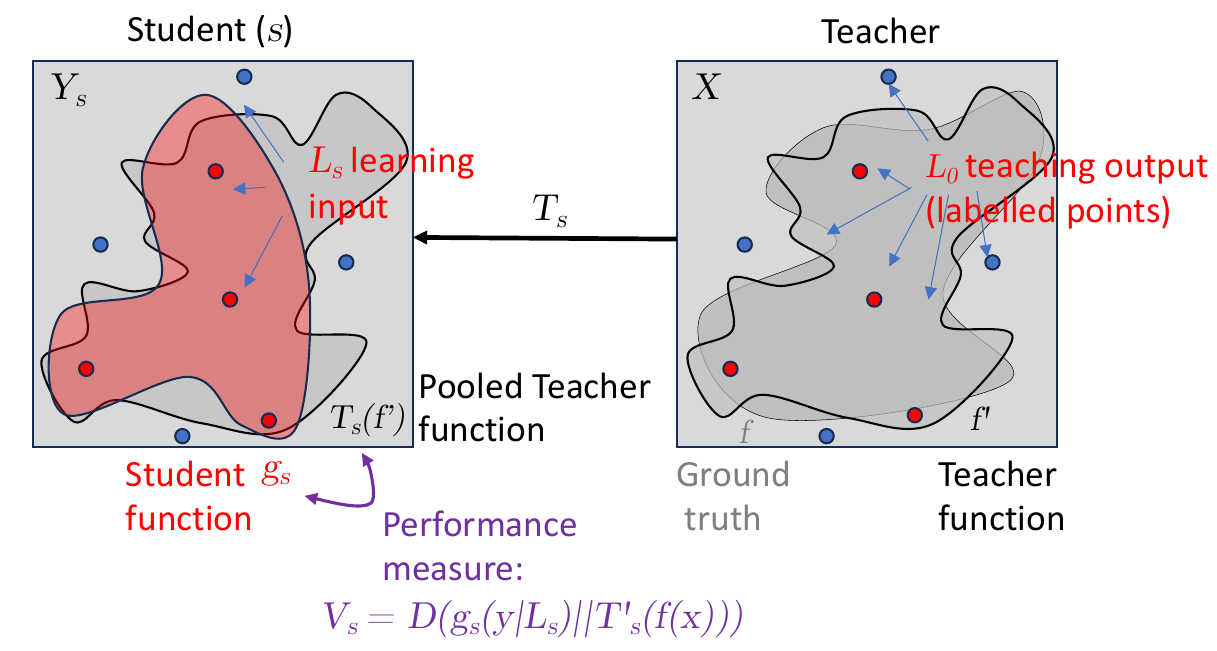}
    \caption{Schematic of teaching and representational alignment. Teachers and students have distinct representational spaces ($X, Y_s$) with some mapping between them ($T_s$). There is a true label function ($f$) that can be projected onto both the teacher and student spaces, but a teacher may not perfectly know this true label function and have their own, diverging label function ($f'$). The teacher designs curricular materials ($L_0$; a set of examples paired with labels) that are projected to each student's space ($L_s$), where each student uses them to learn a label function ($g_s$). Each student's performance ($V_s$) is then measured as the divergence between the learned label function and the hidden true label function ($T'_s(f)$).}
    \label{fig:schematic}
\end{figure}

We offer a deeper theoretical formalism for our setting. Figure \ref{fig:schematic} shows a schematic of our teaching and representation alignment framework. Consider a space $X$ of stimuli. We consider the case where the teacher tries to teach the students some function $f:X\rightarrow C$. We illustrate a simple case in Figure \ref{fig:schematic} wherein $C$ is a binary classification $C=\{0,1\}$ dividing $X$ into two regions ($C=0$ and $C=1$ are represented in Fig. \ref{fig:schematic} in light and dark gray, respectively). The teacher observes label function $f':X\rightarrow C$, which may be different from $f$.   

The teacher chooses $n$ points from the space  ${x_1,x_2,\ldots,x_n} \in X$ and assigns labels to the points $l_i$. The teacher materials can be represented by the labeled points : $L_0=(x_i,l_i)_{i=1,\ldots,n}$. To represent the fact that students' representations may differ from that of the teacher, we assume that the student $s$ has a space $Y_s$ that corresponds to the student's representations. Note that each student is part of some classroom or population of students $s\in \mathcal{S}$.
 
Next, we assume there is some transformation $T_s:X\rightarrow Y_s$. We assume that the function $T_s$ is also selected from some parametric function $T_s \sim \mathcal{T}$. The student $s$ observes stimuli presented by the teacher $y_i=T_s(x_i)$ and labels $l_i$. The student learning input (i.e., the teaching materials mapped into the student's space) is thus $L_s=(T_s(x_i),l_i)_{i=1,\ldots,n}$. From that, the student infers the labeling for the rest of the space, which can be represented as the learning function $g_s(y|L_s)$. The classification performance of the student is tested over additional test points where the expected performance of the student is  $V_s=D(g_s(y|L_s)||T_s(f(x)))$. Here, $D$ represents some distance measure. 

\section{Additional Details on Task Setup for the Single Teacher-Student Setting}

% \section{Additional Results into a Relationship between Representation Alignment and Teaching} 

\subsection{Additional Results on Machines Teaching Humans}

We present the correlations between average huamn and student teacher accuracy in Table ~\ref{tab:human-results}. 

\begin{table}[h!]
\centering
\caption{Pearson correlations (with associated $p$-values) of average human student accuracy and representational alignment of the machine teacher across the various conditions.}
\begin{tabular}{lccc}
\toprule
                         & \textbf{Quadrants} & \textbf{Columns} & \textbf{Both} \\
\midrule
\textbf{simple-features} & 0.91 ($p$=0.013)       & 0.59 ($p$=0.221)     & 0.59 ($p$=0.054) \\
\textbf{salient-dinos}   & 0.52 ($p$=0.286)       & 0.86 ($p$=0.027)     & 0.63 ($p$=0.037) \\
\bottomrule
\end{tabular}

\label{tab:human-results}
\end{table}

\section{Additional Details on 
Computational Simulations}

\subsection{Compute Details}

All experiments were run on an 8-core, 16 GB memory laptop. Experiment were run exclusively on CPUs and were all runnable within at most three hours. 

\begin{figure}[htbp]
    \centering

    \begin{subfigure}
        \centering
        \includegraphics[width=0.45\textwidth]{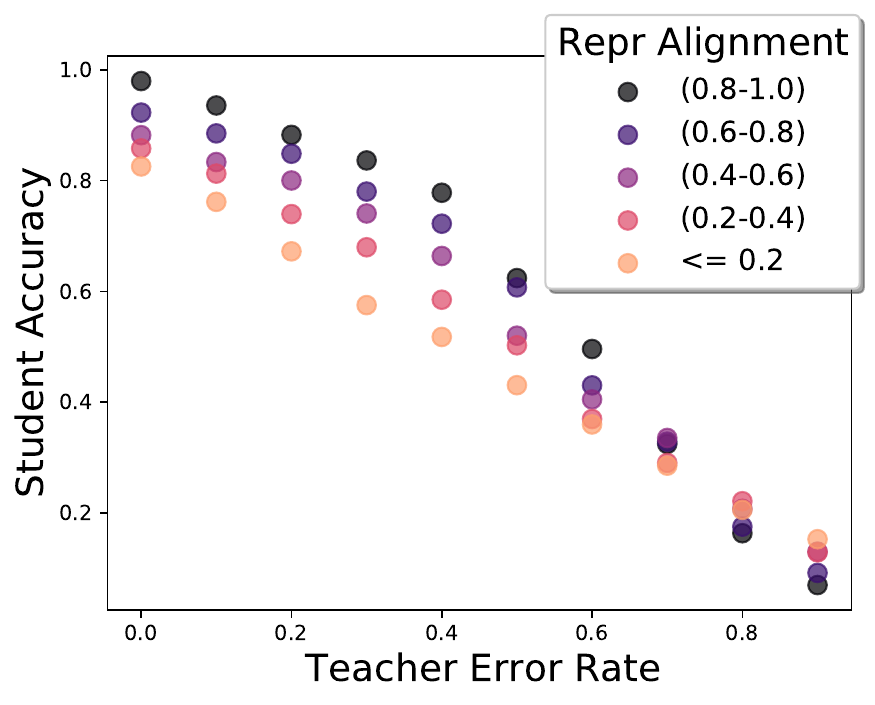} % Replace with your image file
    \end{subfigure}
    \hfill
    \begin{subfigure}
        \centering
        \includegraphics[width=0.45\textwidth]{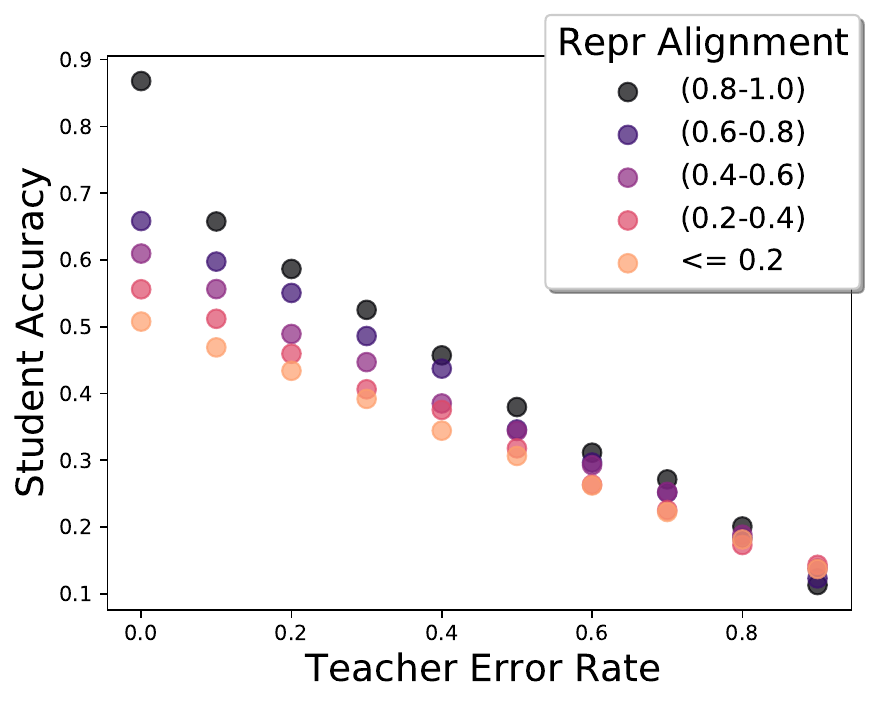} % Replace with your image file
    \end{subfigure}
    \caption{Utility curves on a $6\times6$ grid for different label structures. \textbf{(Left:)} 4 class underlying label-per-quadrant; \textbf{(Right:)} 6 class underlying label-per-column.} 
    \label{fig:diff-structures}
\end{figure}

\subsection{Constructing the Dyadic Utility Curve}
\label{dyadic-utility-curve} 

To construct our utility curve in Section \ref{simulations}, we sweep over a range of possible teacher error rate parameterizations (from $0$ to $0.9$ in increments of $0.1$) and representation corruption levels (from $0$ to $1.0$ in increments of $0.01$). We always use the same ``student'' and corrupt teachers over the respective student grid. We compute the representation alignment between the student and corrupted teacher; as the pairwise swaps (``corruptions'') are randomly made over a fraction of the grid parameterized by the corruption level, we bucketize the resulting observed representation alignment between student and teacher. We then sample $10$ different seeds of selections for each teacher and average student performance. We repeat the same sweeps over teacher parameterizations for our two labeling schemes: grids wherein each column corresponds to one label ($N$ labels for an $N \times N$ grid), and one where each quadrant corresponds to one label (four labels). We average the resulting utility curves across label types. 

\paragraph{Impact of Underlying Label Structure}
We depict the separate utility curves in Figure ~\ref{fig:diff-structures}. Notably, we observe different utility curves for different label structures. While there are some minor rank swaps between teachers across the structures, we see high Spearman rank correlation ($\rho = 0.994, p << 10e-48$) between the two settings underscoring general consistency in teacher orderings.

\subsection{Utility Curve over Classrooms}
\label{classroom-utility-curve}

The utility curves that we construct in Section \ref{simulations} and \ref{dyadic-utility-curve} were always constructed with respect to a single student (in a respective, ``dyadic''\footnote{Pairing two agents -- one student and one teacher.}). However, in our classroom settings, we also corrupt the students' representations to simulate representational diversity. We sample a new utility curve, wherein, for each teacher parameterization (same error rate parameterization as above, with representation corruptions now in increments of $0.1$), we sample $10$ different student corruptions ranging over $0$ to $0.9$ in increments of $0.1$. We build this curve only for the column label type. We then bucketize the teacher error rate as well as the representation alignment such that we can index into the curve to extract an ``expected average performance'' for any student-teacher pair.

%(with the same settings as our dyadic grid, but coarser representation align

\subsection{Constructing Classroom Pools} 
\label{classroom-pools}

We construct two different classroom pools in Section \ref{classroom-matching}: unstructured and structured. Here, we provide additional details on how we sampled students and teachers for each pool type.

\paragraph{Unstructured Pool} All students and teachers are sampled independently. We sample $1000$ students and $30$ teachers with corruption levels (pairwise swaps) sampled from a beta distribution ($\alpha=1.5, \beta=2.5$) to ensure that we have some students that are reasonably aligned. We sample teacher error rate uniformly over the range $0-0.5$. %We include the resulting distribution on representation alignment and teacher error rates in Figures [...] respectively.  

\paragraph{Structured Pool} In the structured setting, we construct \textit{clusters} of similar students and teachers. We prespecify a number of clusters $M$ and number of students per cluster. Clusters are designed to span a range of levels of representation alignment over the ``original'' grid. We loop over possible representation alignments corruptions ranging from $0$ to $1$ in increments of $1/M$. For each cluster, we sample a ``seed'' student using that corruption level. We then sample [notation?] students ontop of this cluster with a representation corruption of $0.01$ ontop of the base student to ensure students share similar (but some variation) in their representation. For each cluster, we sample a teacher with error rate uniformly from $0-0.5$ and representation with a similar slight possible corruption (sampled uniformly from $0-0.01$) ontop of the seed student, thereby ensuring that there \textit{would} be a representationally similar teacher for each student in each cluster if provided. However, to simulate gaps in coverage of particular representation characterezations, we randomly drop some teachers from the pool. 

% \subsection{Impact of Number of Clusters} 
% \label{num-clusters}

% We investigate the impact of the number of clusters and proportion of available students on teaching outcomes in Figure \ref{} [todo]. 

\subsection{Additional Classroom Matching Results}

\begin{figure}
    \centering
    \includegraphics[width=0.49\linewidth]{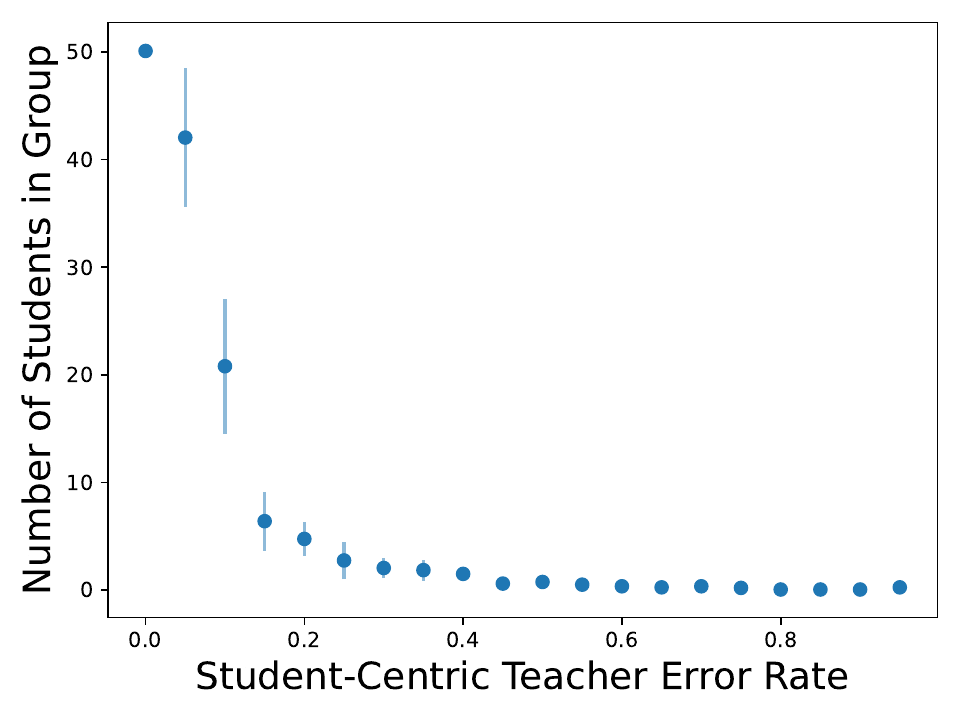}
    \includegraphics[width=0.49\linewidth]{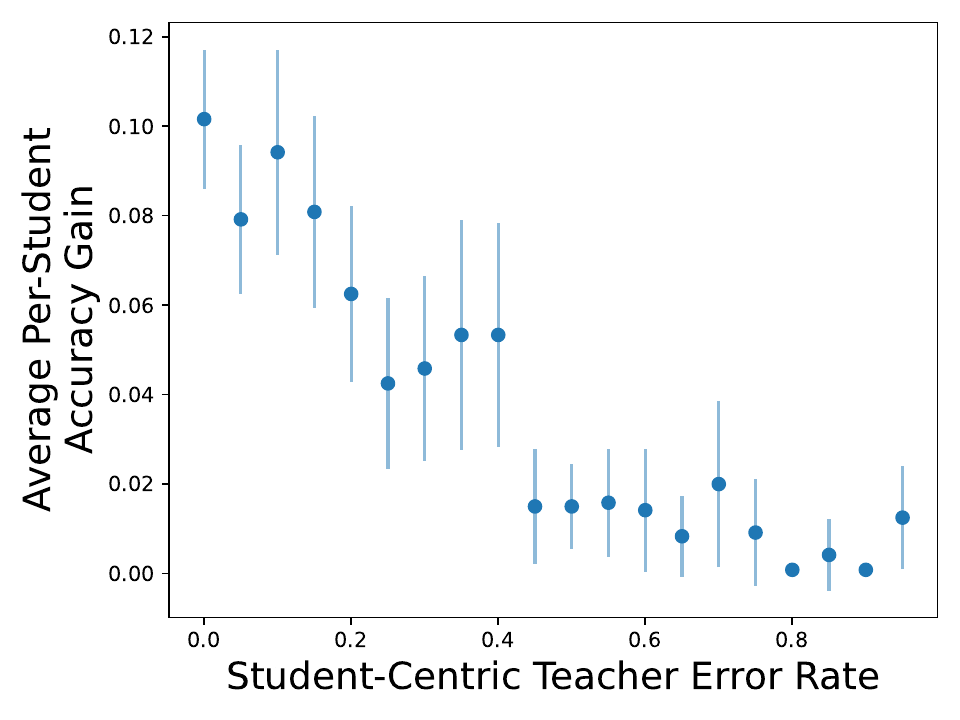}
    \caption{\textbf{(Left:)} Group sizes from greedily incorporating the lowest performing students into the classroom of a single student-centric teacher. \textbf{(Right:)} Average accuracy gains (out of 1.0) in performance for students grouped with the student-centered teacher, on top of what they would have achieved from a self-centered teacher. Error bars are standard errors over $20$ seeds of student-centric teacher groupings for a sampled structured pool of students and teachers.}
    \label{fig:student-centric-greedy}
\end{figure}

We include additional results into classroom matching in Table ~\ref{tab:classroom-matching} and a relationship between group size and learning outcomes in Figure ~\ref{fig:student-centric-greedy}. 

\begin{table}[]
\caption{Student learning outcomes (accuracy) from different classroom matching approaches in the unstructured pool setting. We compute the average student performance across all $N=1000$ pooled students (paired with potentially $M=30$ teachers), as well as accuracy over the bottom and top 10\% of students in each matching, respectively. We additionally compute the proportion of students who achieve ``passing'' marks (set to a moderately high threshold of 45\% accurate, given chance guessing is 16.6\% on our 6x6 grid). Higher is better for all metrics. $\pm$ indicates standard errors computed over 10 sampled pools and associated assignments.}
\centering
\begin{tabular}{@{}lllll@{}}
\toprule
Method  & Avg Acc & Bottom 10\% & Top 10\% & Pass Rate \\ 
\midrule
Random  & 0.32 $\pm$ 0.01 & 0.17 $\pm$ 0.01 & 0.51 $\pm$ 0.01 & 0.08 $\pm$ 0.01 \\
MOOC    & 0.38 $\pm$ 0.01 & \textbf{0.25 $\pm$ 0.00} & 0.54 $\pm$ 0.04 & 0.18 $\pm$ 0.05 \\
Ours    &  \textbf{0.39 $\pm$ 0.01} &  0.24 $\pm$ 0.01 & \textbf{0.59 $\pm$ 0.04} & \textbf{0.23 $\pm$ 0.03} \\
\hline
Optimal & 0.49 $\pm$ 0.01 & 0.36 $\pm$ 0.00 & 0.70 $\pm$ 0.03 & 0.53 $\pm$ 0.01 \\
\bottomrule
\end{tabular}
 %Classroom matching, based on our utility curve. 1000 students, 30 teachers. 6x6 simple grid. MOOC is giving all students the lowest error rate teacher. Pass rate set to 0.45. Error bars depict standard error. \textit{caption in progress}}
\label{tab:classroom-matching}
\end{table}

\subsection{Generalization to the Dino Stimuli}
\label{dinos-matching generalization}

We explore generalization of our utility curve constructed in the simple-features setting to our salient-dino stimuli. We repeat our two different pool types, which we depict in Tables \ref{tab:classroom-matching-dino} and \ref{tab:classroom-matching-clustered-dino}, respectively. We find that our utility curves generalize nicely to different grid sizes and stimuli type, yielding student outcomes that on average appear to boost student accuracy particularly for the students in the top-performing group than baselines which do not account for representation misalignment (MOOC).

\begin{table}[]
\centering
\caption{Student learning outcomes (accuracy) from different classroom matching approaches in the unstructured pool setting for the dino stimuli. We again compute the student performance across all $N=1000$ pooled students (paired with potentially 30 teachers). Error bars are again computed over 10 different sampled pools.} 
\begin{tabular}{@{}lllll@{}}
\toprule
Method  & Avg Acc & Bottom 10\% & Top 10\% & Pass Rate \\ 
\midrule
Random  &  0.29 $\pm$ 0.00 & 0.16 $\pm$ 0.01 & 0.46 $\pm$ 0.01 & 0.04 $\pm$ 0.01 \\
MOOC    & 0.34 $\pm$ 0.02 & \textbf{0.22 $\pm$ 0.00} & 0.51 $\pm$ 0.07 & 0.09 $\pm$ 0.05 \\
Ours    &  \textbf{0.36 $\pm$ 0.01} &  \textbf{0.22 $\pm$ 0.00} &  \textbf{0.64 $\pm$ 0.08} & \textbf{0.15 $\pm$ 0.03} \\
\hline
Optimal & 0.44 $\pm$ 0.01 & 0.31 $\pm$ 0.00 & 0.69 $\pm$ 0.08 & 0.32 $\pm$ 0.02 \\
\bottomrule
\end{tabular}

\label{tab:classroom-matching-dino}
\end{table}

\begin{table}[]
\centering
\caption{Student learning outcomes (accuracy) from different classroom matching approaches in the structured (clustered) pool setting for the dino stimuli. We again have 10 representationally distinct clusters, each with 50 students, and sample 5 available teachers across the clusters.}
\begin{tabular}{@{}lcccc@{}}
\toprule
Method  & Avg Acc & Bottom 10\% & Top 10\% & Pass Rate \\ 
\midrule
Random  & 0.29 $\pm$ 0.02 & 0.18 $\pm$ 0.02 & 0.44 $\pm$ 0.03 & 0.06 $\pm$ 0.03 \\
MOOC    & 0.34 $\pm$ 0.03 & \textbf{0.23 $\pm$ 0.02} & 0.50 $\pm$ 0.08 &  0.10 $\pm$ 0.06 \\
Ours    & \textbf{0.35 $\pm$ 0.03} & \textbf{0.23 $\pm$ 0.02} & \textbf{0.54 $\pm$ 0.08} & \textbf{0.13 $\pm$ 0.06} \\
\hline
Optimal & 0.38 $\pm$ 0.02 & 0.28 $\pm$ 0.01 & 0.56 $\pm$ 0.07 &  0.15 $\pm$ 0.06 \\
\bottomrule
\end{tabular}

\label{tab:classroom-matching-clustered-dino}
\end{table}

\subsection{Additional Details on Student-Centric Teacher}
\label{student-centric-teacher-details}

In contrast to our self-centered teacher, our student-centric teacher does not use its own representation to select examples to provide to the student. Instead, the student-centric teacher is endowed with an \textit{inner optimization loop} over the students assigned to it, whereby the teacher loops $T$ times over the students and randomly selects one point per category (using the teacher's believed class -- the teacher may not know the true categories) and measures the expected performance of each student if that set of examples were revealed. Note, the teacher computes the expected accuracy of each student using against its belief of the true categorization (which may be incorrect). The teacher then chooses the set of examples that attains the highest average accuracy over students. Here, we set $T$ to 100; exploring the impact of varied $T$ is a sensible next step. Exploration of alternate optimization functions, e.g., optimizing over the minimum attained performance over the students in the teacher's classroom rather than average classroom performance are also ripe ground for future work. %and importantly intersect with conversations around fairness.

\subsection{Relationship Between Classroom Size and Student-Centric Teacher Utility} 
\label{cls-size-student-centric}

\begin{figure}
    \centering
    \includegraphics[width=0.8\textwidth]{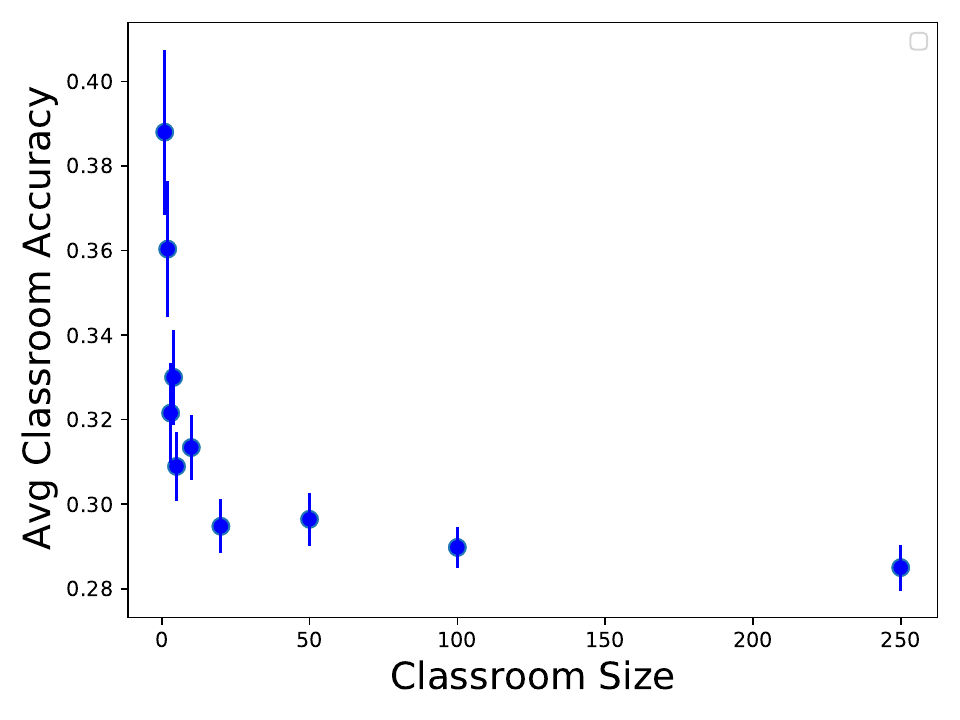}
    \caption{Average performance of students paired with a student-centric teacher as a function of classroom size. Error bars depict standard error over runs.}%PLACEHOLDER -- formatting and final run in progress. This gives a sense of the relationship though between classroom size and average student performance in the classroom.}
    \label{fig:cls-size-perf}
\end{figure}

We conduct a deeper dive into the relationship between classroom size and student performance by sampling sweeping over a range of teacher error rates (between $0$ and $0.5$ in increments of $0.1$) and classroom sizes. We sweep over 10 samples for each setting, where each self-centered teacher optimizes for their pool through $T=100$ inner loop iterations. Students are drawn from our structured, clustered pool. We then marginalize over our sampled error rates to compute an expected average classroom accuracy per classroom size. We find that the performance of students in a classroom with a student-centric teacher falls off rapidly as a function of classroom size and then plateaus in Figure \ref{fig:cls-size-perf}.

\section{Additional Human Experiment Details}
\label{human-exp-details-appendix}

\subsection{Participant Recruitment and Compensation}
Participants were recruited from Prolific and were paid \$12/hr plus a 10\% bonus if they responded reasonably (i.e., did not select labels randomly or choose the same label for all stimuli). The research did not contain risks to participants, and they were able to opt out at any time. The institution of the principal investigator obtained IRB approval for this experiment, and participants gave informed consent under this protocol.

\subsection{Task Instructions}

\begin{figure}[htbp]
    \centering
    \includegraphics[width=0.90\textwidth]{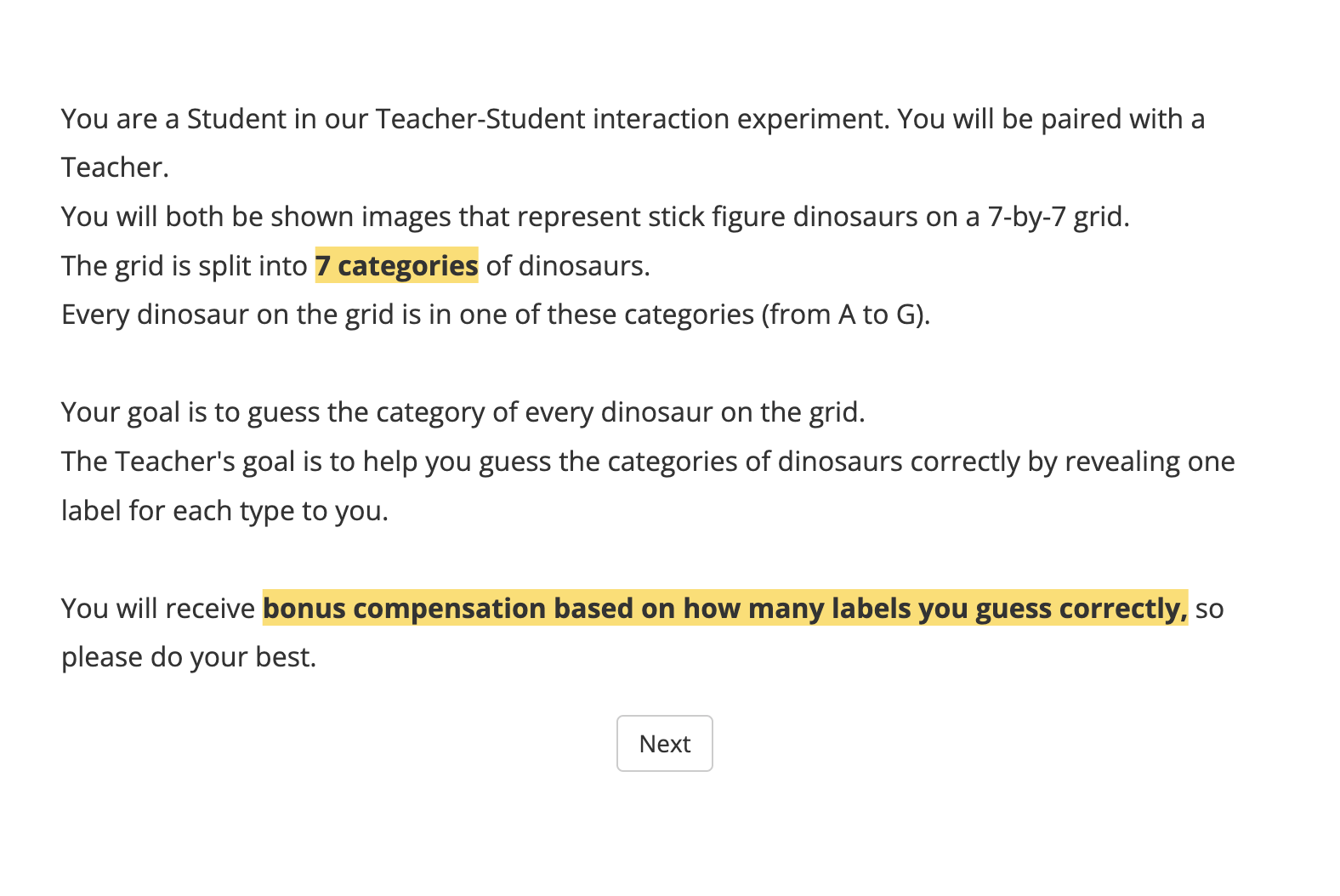}
    \caption{Experiment instructions displayed to all participants, introduced paragraph by paragraph. The only changes to instructions were to modify the type of stimuli (``empty cells'', ``images that represent stick figure dinosaurs''), size of the grid ($6\times6$, $7\times7$), the number and names of categories (4; A-D or 6/7; A to F/G).}
    \label{fig:experiment-instructions}
\end{figure}

\begin{figure}[hbp]
    \begin{subfigure}
        \centering
        \includegraphics[width=0.65\textwidth]{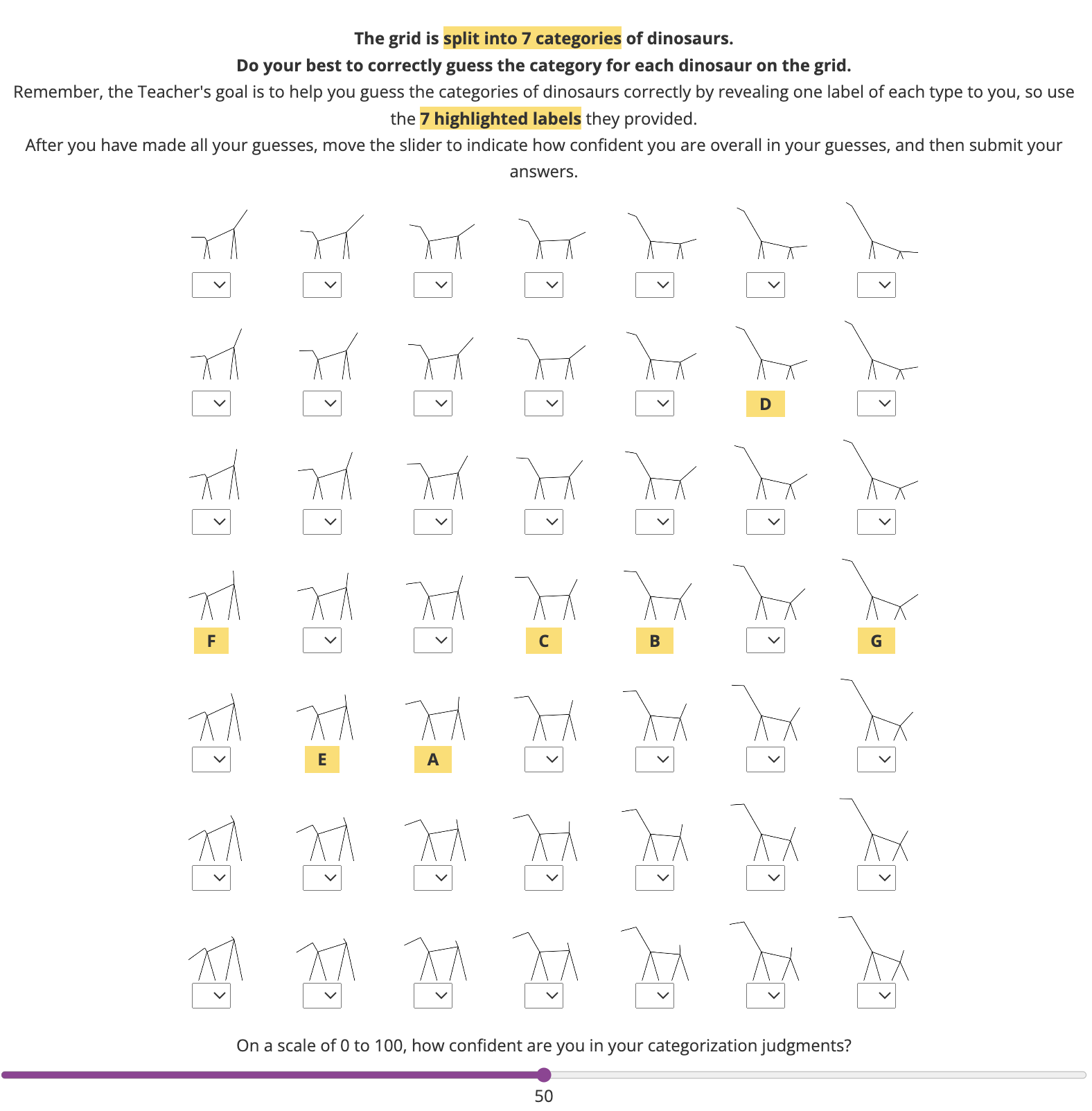}
    \end{subfigure}
    \hfill
    \begin{subfigure}
        \centering
        \includegraphics[width=0.3\textwidth]{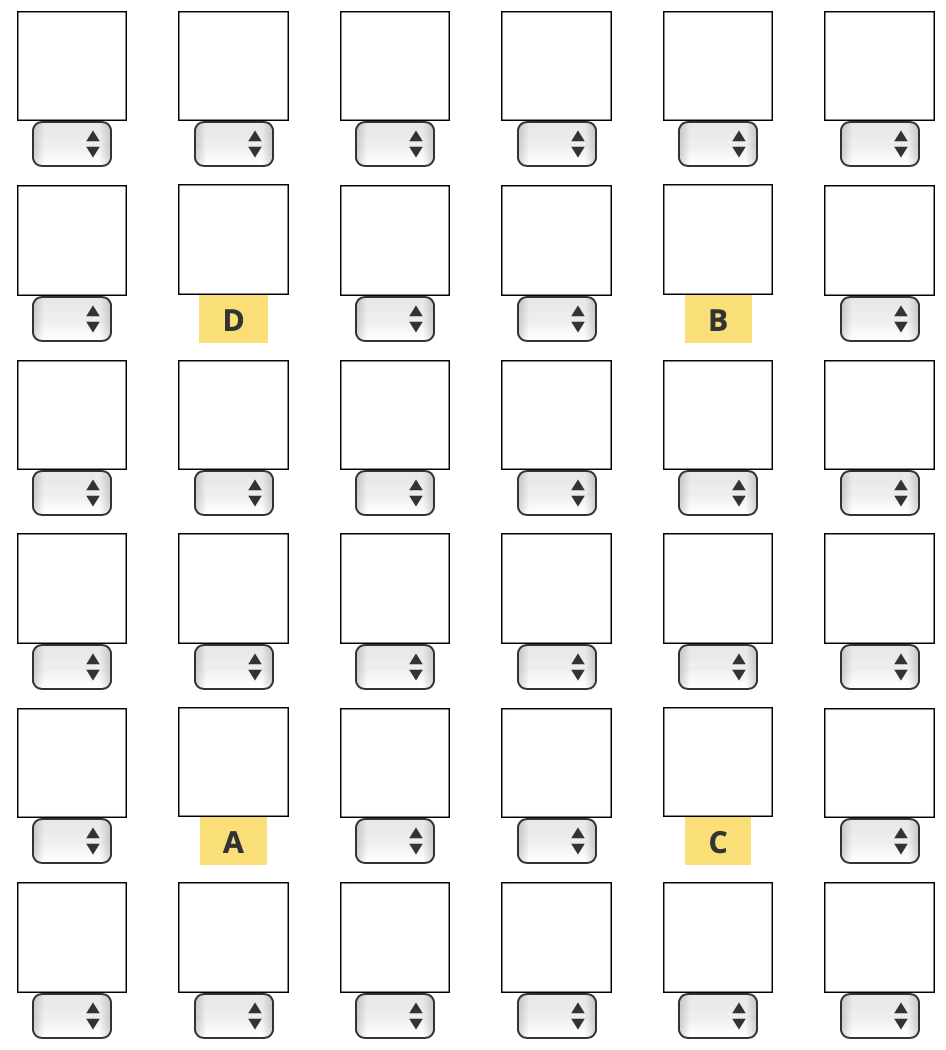}
    \end{subfigure}
    \caption{Above are two example views of the experiment. All participants, after viewing the instructions in Figure \ref{fig:experiment-instructions} were taken to a page that contained a grid and the labeled stimuli. They were asked to categorize stimuli via a dropdown menu selection. Finally, they rated their confidence using a scale below the stimulus grid. \textbf{Left:} \textit{salient-dinos}, 7 (``col'') categories, medium-alignment teacher. \textbf{Right:} \textit{simple-features}, 4 (``quad'') categories, high-alignment teacher.}
    \label{fig:experiment-dino-grid}
\end{figure}

We include the full set of instructions provided to participants in Figure \ref{fig:experiment-instructions} and sample interfaces in Figure \ref{fig:experiment-dino-grid}. 

\subsection{Further Analyses} 
\label{further-analyses}
\begin{figure}
    \centering
    \includegraphics[width=0.49\linewidth]{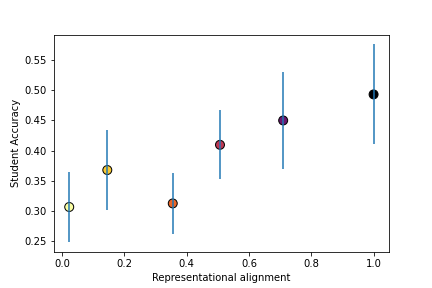}
    \includegraphics[width=0.49\linewidth]{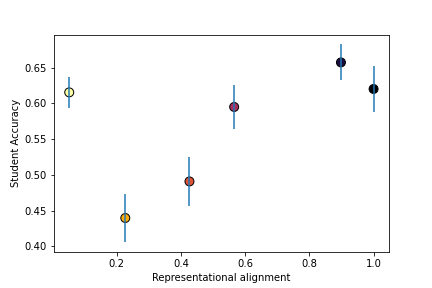}
    \includegraphics[width=0.49\linewidth]{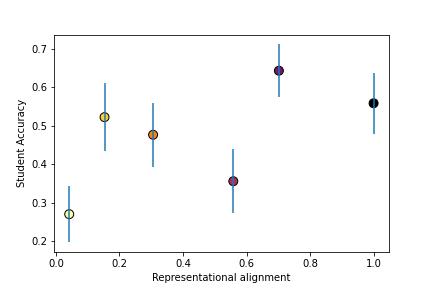}
    \includegraphics[width=0.49\linewidth]{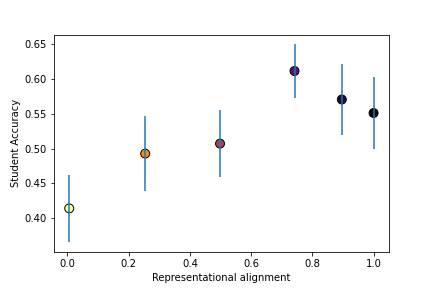}
    \includegraphics[width=0.49\linewidth]{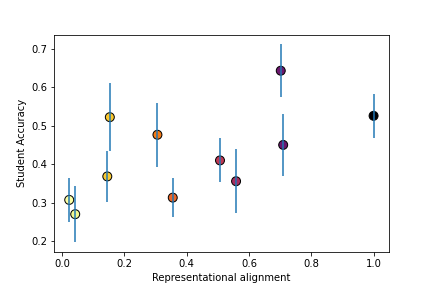}
    \includegraphics[width=0.49\linewidth]{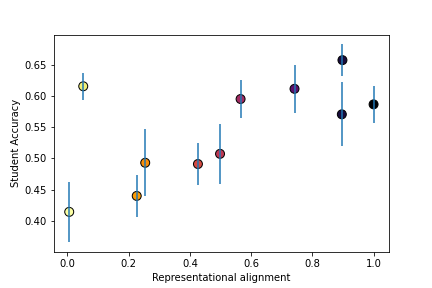}
    \caption{Average human student classification accuracy at various levels of representational alignment. Error bars correspond to one standard error. \textbf{(Left:)} Results from simple-features setting. \textbf{(Right:)} Results from salient-dinos setting.
    \textbf{(Top:)} One class per quadrant. \textbf{(Middle:)} One class per column (6 for simple-features, 7 for salient-dinos).
    \textbf{(Bottom:)} Combined results.}
    \label{fig:zero-error}
\end{figure}

\paragraph{Simulating teacher error in human experiments}
All human experiments were run with machine teachers set to zero error, as collecting all combinations of teacher error and representational alignment would be prohibitively expensive. Instead, we simulate the effect of teacher error in a post-hoc analysis by corrupting the true underlying labels in the same way we corrupted the teacher labels for the simulation experiments (i.e., error rate corresponds to the probability with which we flip each true label to be a different label). Human student accuracy was then recomputed against these corrupted true labels. The original human student results with no simulated teacher error are reported in Figure~\ref{fig:zero-error}.

\end{document}